# Uncertainty-Aware Model-Based Reinforcement Learning with Application to Autonomous Driving

Jingda Wu, *Student Member, IEEE*, Zhiyu Huang, *Student Member, IEEE*, and Chen Lv, *Senior Member, IEEE*

*Abstract*— To further improve the learning efficiency and performance of the reinforcement learning (RL), in this paper we propose a novel uncertainty-aware model-based RL (UA-MBRL) framework, and then implement and validate it in autonomous driving under various task scenarios. First, an action-conditioned ensemble model with the ability of uncertainty assessment is established as the virtual environment model. Then, a novel uncertainty-aware model-based RL framework is developed based on the adaptive truncation approach, providing virtual interactions between the agent and environment model, and improving RL's training efficiency and performance. The developed algorithms are then implemented in end-to-end autonomous vehicle control tasks, validated and compared with state-of-the-art methods under various driving scenarios. The validation results suggest that the proposed UA-MBRL method surpasses the existing model-based and model-free RL approaches, in terms of learning efficiency and achieved performance. The results also demonstrate the good ability of the proposed method with respect to the adaptiveness and robustness, under various autonomous driving scenarios.

## I. INTRODUCTION

As a crucial component of the future mobility systems, autonomous vehicles (AVs) have been attracting great attentions from both academia and industry. Developing safe, smart and smooth autonomous driving strategy is of great importance to promote the application and deployment of AVs in real world. Comparing to complicated rule-based strategies, learning-based autonomous driving approaches become more advantageous in dealing with highly dynamic driving tasks, due to their powerful nonlinear reasoning ability [1]. Among learning-based approaches, reinforcement learning (RL) is very promising for solve complex autonomous driving problems [2]. The distinguished offline optimization ability of RL rescues developers from heavy-duty algorithm designs, and in the meantime, its computational efficiency also relieves the harsh requirements on on-board computing devices in real-time applications [3].

For RL methods, the optimized policy is learnt from the interactive data between the environment and RL agent [4]. Currently, the representative RL methods include but are not limited to deep Q-learning [5], proximal policy optimization[6], and the soft actor-critic [7]. Substantial RL-based applications in AVs have been reported, and the main tasks include the optimization of intelligent traffic flows [8], decision making and control of autonomous vehicles [9, 10]. Although these works showed the great post-training ability of RL in real-time applications, they also revealed a crucial drawback that the offline training is time consuming and requires heavily on the computation resources [11]. A reason behind is that the interactions between the agent and environment are inefficient, and every single iteration can only generate limited data [12].

To mitigate the above problem, the concept of model-based RL (MBRL) was investigated. Using data-driven technologies, the characteristics, behaviors, and relations in the actual environment can be represented holistically to create high-fidelity virtual models. Besides the actual interactions between the agent and the actual environment, the virtual environment model, which can predict the state transition under agent's actions, is expected to be a useful complement to generate additional data and accelerate the training. In this context, recently the MBRL attracted intensive research interests from both theoretical and application aspects [12,13].

Model-based approaches can be divided into four types based on their different usage of the model [14]. The first type is the Dyna-style methods. They leverage the virtual interactions between the RL agent and the virtual environment and feed the generated data to the experience replay buffer for RL optimization [15-17]. Compared to the conventional RL methods, the efficiency can be improved due to the virtual interactions conducted within the model. For the second type, approaches incorporating model predictive control (MPC) have been presented [18]. In these methods, an MPC controller optimizes the RL agent's behavior policy by predicting and

J. Wu, Z. Huang, and C. Lv are with the School of Mechanical and Aerospace Engineering, Nanyang Technological University, Singapore, 639798. (e-mails: {jingda001,zhiyu001}@e.ntu.edu.sg, lyuchen@ntu.edu.sg)

Corresponding author: C. Lv.

This work was supported in part by A*STAR National Robotics Programme (No. SERC 1922500046), the A*STAR AME Young Individual Research Grant (No. A2084c0156), and the Alibaba Group through Alibaba Innovative Research (AIR) Program and Alibaba-NTU Singapore Joint Research Institute (JRI) (No. ANGC-2020-012).

planning ahead within the virtual environment at each training step. The third type is the value augmentation scheme, which uses the rollout data generated from the virtual environment to construct an intensified value function [19-21]. By facilitating the convergence of value function, the training process of the RLs is expected to be accelerated. The fourth type is the analytic-gradient scheme. It constructs a differentiable model to optimize the RL's policy function through the gradient metric, which is the gradient of the model's feedback with respect to the policy function [14, 22].

However, in the aforementioned model-based RL methods, the model accuracy is currently a bottleneck affecting their performance [23]. Despite the validated feasibility in simple environments, most of them can be hardly transplanted directly to real world tasks with complex multi-agent interactions and uncertainties. Since the dynamic driving environment is particularly difficult to be modeled accurately, it is challenging to develop autonomous driving strategy using MBRL-based methods. To solve the above issue of environment modelling, a smart trade-off between the utilization extent and accuracy of the model should be made. Within the existing model-based RL methods, analytic-gradient approaches rely fully on the imaginary environment and suffer from severe issues caused by model sensitivity [24]. Value augmentation-based solutions could face the importance sampling problem when updating the value function [3], limiting its application to off-policy RLs. And MPC-based methods impose heavy computation burden, which is not suitable for exploration-based optimization. Hence, to develop model-based RL for autonomous driving, the Dyna-style approach is a promising candidate.

In this paper, we propose a Dyna-style model-based RL framework with its application to end-to-end autonomous driving. More specifically, four main contributions are introduced as follows. First, a virtual-model-enabled actor-critic RL framework is proposed for improving the training efficiency and performance. Second, a novel environment model is established to predict the transition dynamics with the ability of uncertainty assessment. Third, a new model-based entropy maximized RL algorithm, in which the model uncertainty is considered throughout the training process to realize adaptive utilization of the model prediction, is proposed. Lastly, the developed uncertainty-aware model-based RL algorithm is implemented in end-to-end autonomous driving tasks, and its performance is validated across varying testing scenarios.

The remainder of this paper is organized as follows. The high-level uncertainty-aware model-based RL framework is introduced in Section II. Then, the environment modelling approach is proposed in Section III. Under the developed framework, the model-based actor-critic algorithm is developed based on the uncertainty-aware environment model in Section IV. Section V introduces the implementation details of the developed algorithms in autonomous driving tasks. Section VI provides the validation, results, and discussions. Lastly, conclusions are drawn in Section VII.

## II. THE UNCERTAINTY-AWARE MODEL-BASED REINFORCEMENT LEARNING FRAMEWORK

In this section, the high-level framework of the proposed uncertainty-aware model-based RL for autonomous driving is introduced. As illustrated in Fig. 1, the actor-critic RL agent interacts with the actual driving environment. The generated actual interaction data is stored into the experience buffer $\mathcal{D}_E$, which is leveraged in the conventional RL for improving the learning policy. Besides, in our proposed framework, an additional virtual environment model is uniquely designed and evolved using the actual experience buffer to construct the transition dynamics of the driving environment. Leveraging this virtual model, the agent can then interact with the virtual environment, in parallel to the actual interactions. In this way, we are able to generate more virtual experience $\mathcal{D}_M$ via predictions in the environment model, and therefore expedite the RL's learning process with less actual interactions and iterations.

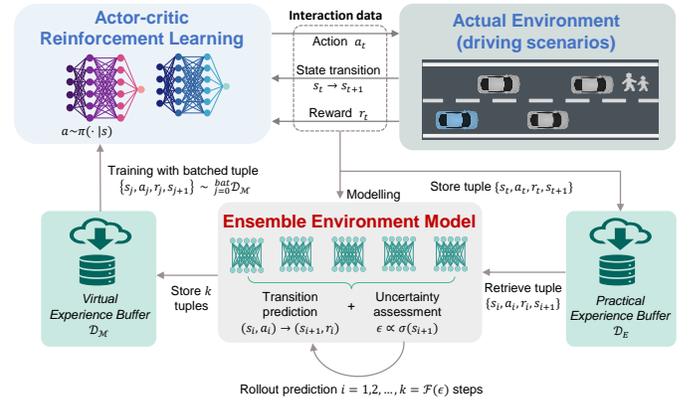

Fig. 1. The high-level architecture of the proposed UA-MBRL method with application in autonomous driving task.

More specifically, the virtual environment model starts its prediction from an actual state and conducts a consecutive rollout process. Consisted of multiple randomly initialized entities, the ensembled model can assess the uncertainty of the output. Thus, one expected innovation in this developed framework is that the number of rollout steps and unreliable predicted experience can be adaptively determined and truncated based on the assessed model uncertainty. Aiming at autonomous driving tasks, various traffic scenarios will be designed for RL's training and testing for validating the proposed method. The RL agent is expected to manipulate the ego vehicle in an end-to-end manner and conduct the lateral motion control without violating designed constraints. The training efficiency and practical execution performance of the RL agent in autonomous driving scenarios will be studied in this work.

## III. VIRTUAL ENVIRONMENT MODELLING WITH UNCERTAINTY ASSESSMENT

Under the developed novel RL framework, first, an efficient model of the environment should be established for realizing rapid and uncertainty-aware prediction of the agent-environment virtual interactions.

A parameterized model $\mathcal{M}$ with parameter $\psi$ is established to predict the transition dynamics in the autonomous driving scenarios. This expected function can be realized through an action-conditioned predictive model [25]. With the encoder-decoder framework, the environment model receives compound state variable and prospective action, and outputs the predicted next-step state as well as the reward, respectively. This mapping relationship can be expressed as:

$$\mathbf{s}_{t+1}, r_t \leftarrow \mathcal{M}_\psi(\mathbf{s}_t, \mathbf{a}_t) \qquad (1)$$

where $\mathbf{s}_t$ denotes the input state, $\mathbf{a}_t$ is the input action, $r_t$ is the reward, and $\mathbf{s}_{t+1}$ is the predicted next-step state.

To achieve a successful prediction, it is important to provide sufficient state information to the environment model. In the autonomous driving tasks, the ego vehicle's motion and the surrounding situations can be described by a compound state space of the visual and physical information. Specifically, the visual states can be obtained from the processed semantic image, and the physical states include the yaw angle, lateral position, and lateral velocity of the ego vehicle. Then the environment model can be developed using a deep neural network, and its structure is illustrated in Fig. 2. Some additional information is provided in the Appendix Table I.

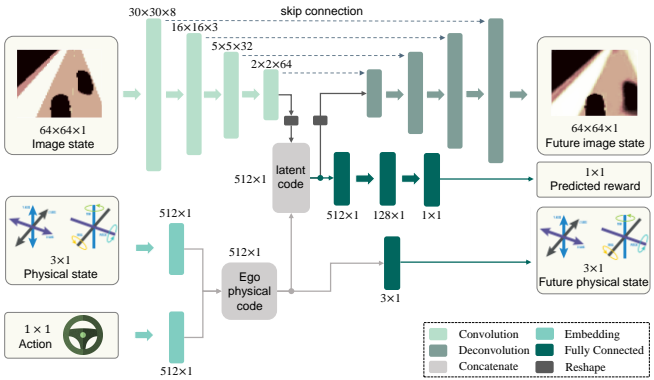

Fig. 2. Schematic diagram of the virtual environment model.

Since physical states have explicit meanings and are only associated with the ego vehicle, thus they are relatively easier to be predicted. However, the prediction of image states is more challenging, because it requires a good high dimensional representation as well as additional inference of the feedback states of the traffic participants. Therefore, in our model, the physical states are utilized to achieve a posterior for the image state prediction. As shown in the lower part of Fig. 2, the input action is combined with the ego vehicle's physical states for predicting the physical variables at the next step. Accordingly, the intermediate variable, i.e., the obtained ego vehicle's physical code, can be taken as posterior information of the state transition. Thus, we send it to the upper module for enhancing the image prediction ability.

To capture the aleatoric uncertainty of the model, we assume the outputs subject to Gaussian distribution $\mathcal{N}$ with diagonal covariances [26]. Thus, the output of the model is expressed as:

$$\mathcal{M}_\psi = \mu_\psi, \Sigma_\psi \qquad (2)$$

where $\mu$ and $\Sigma$ represent the mean and covariance of the Gaussian distribution, respectively.

Then the transition dynamics is shown as:

$$\Pr(\cdot \mid \mathbf{s}_t, \mathbf{a}_t) = \mathcal{N}\left(\mu_{\psi_m}(\mathbf{s}_t, \mathbf{a}_t), \Sigma_{\psi_m}(\mathbf{s}_t, \mathbf{a}_t)\right) \qquad (3)$$

where $\Pr(\cdot)$ represents the probabilistic transition, and subscript $m$ denotes the "next-step state" or the "reward".

In the training process of the environment model, we let the model to output the mean value only and adopt a fixed value as the covariance. In this way, the objective function of the environment model can be simplified. The mean squared loss function is adopted and expressed as:

$$\mathcal{L}_\mathcal{M}(\psi_m) = \mathop{\mathbb{E}}_{(\mathbf{s}_t, \mathbf{a}_t) \sim \mathcal{D}_E} \left[ \| \mu_{\psi_m}(\mathbf{s}_t, \mathbf{a}_t) - d_{m,t} \|_2^2 \right] \qquad (4)$$

where $\mathcal{D}_E$ represent the experience buffer of the training data, and $d_m$ represents the output value of the model, i.e., the next-step state and reward, respectively.

For the execution of model prediction, the output is sampled from the distribution with fixed variance, where the mean value is varied with the inputs.

In addition to the aleatoric uncertainty, epistemic uncertainty is also crucial for characterizing the model accuracy [26]. Concretely, when the data distribution is shifted due to the interactions between the ever-updated RL policy and the environment, the constructed model would generate inaccurate prediction if encountering untrained inputs, which induces the increasing epistemic uncertainty.

To quantify the epistemic uncertainty, in this paper the ensemble approach is utilized. Five independent and identically distributed models are initialized with different random seeds and trained jointly, which can be expressed as:

$$\mathcal{M}_{ens} = \{\mathcal{M}_\psi^1, \ldots, \mathcal{M}_\psi^N\} \qquad (5)$$

where $N = 5$ represents the ensemble size.

Through uniform sampling, the prediction data can be generated using an arbitrary model from the ensemble set, which skirts the biased prediction.

## IV. UNCERTAINTY-AWARE MODEL-BASED REINFORCEMENT LEARNING ALGORITHM

Based on the above virtual environment modelling, in this section, a novel uncertainty-aware model-based reinforcement learning algorithm is proposed to improve the convergence efficiency and asymptotic performance. First, the base algorithm adopted is introduced. Then, the theoretical boundary of MBRL for effective optimization is discussed. Based on the above two points, the key innovation of the proposed method is elaborated in the last part of this section.

### A. The Soft Actor Critic Algorithm

Because of the validated strong robustness under various dynamic environment, in this work, Soft Actor Critic (SAC) is utilized as the base algorithm for developing the proposed new RL architecture [7, 27].

Within the SAC algorithm, the interaction between the agent and the environment can be formulated as a Markov Decision Process, denoted as $\mathcal{M} = (\mathcal{S}, \mathcal{A}, \mathcal{T}, r, \tau)$. Specifically, $\mathcal{S}$ is the state space consisting of state variable $\mathbf{s}$, $\mathcal{A}$ is the action space which describes candidate actions $\mathbf{a}$, $\mathcal{T}(\mathbf{s}_{t+1}|\mathbf{s}_t, \mathbf{a}_t)$ denotes the state transition of the environment dynamics, $r: \mathcal{S} \times \mathcal{A} \rightarrow \mathbb{R}$ is the reward function, and $\tau = \{\mathbf{s}_t, \mathbf{a}_t, \mathbf{s}_{t+1}, \mathbf{a}_{t+1}, ...\}$ denotes the trajectory under a specific policy.

The ultimate goal of the SAC-RL algorithm is to achieve the action that can maximize the cumulated reward at any given state through encouraged explorations. To this regard, the SAC pursuits the optimal policy $\pi^*$, which can be expressed as:

$$\pi^* = \arg\max_{\pi} \mathbb{E}_{\tau \sim \mathcal{T}, \pi}\left[\sum_{t=0}^{\infty} \gamma^t \left(r(\mathbf{s}_t, \mathbf{a}_t) + \alpha H(\pi(\cdot|\mathbf{s}_t))\right)\right] \quad (6)$$

where $\gamma$ is the discount factor, $H(\pi(\cdot))$ denotes the entropy of the distribution of the policy $\pi$, and $\alpha$ is the temperature factor to balance the pursued high reward and encouraged exploration.

A critic function, i.e. the so-called soft-Q function, is established to represent the above value, which is expressed as:

$$Q(\mathbf{s}_t, \mathbf{a}_t) = \mathbb{E}_{\tau \sim \mathcal{T}, \pi}[(r(\mathbf{s}_t, \mathbf{a}_t) + \gamma \mathbb{E}_{\mathbf{s}_{t+1} \sim \mathcal{T}(\cdot|\mathbf{s}_t, \mathbf{a}_t)}[V(\mathbf{s}_{t+1})]) \quad (7)$$

where $V$ is the state value function that can be calculated as:

$$V(\mathbf{s}_t) = \mathbb{E}_{\mathbf{a}_t \sim \pi}[Q(\mathbf{s}_t, \mathbf{a}_t) - \alpha \log \pi(\mathbf{a}_t|\mathbf{s}_t)]. \quad (8)$$

With the approximation function of deep neural networks, the objective of the soft-Q function $Q_\theta$ can be calculated by minimizing the Bellman residual:

$$\mathcal{L}_Q(\theta_i) = \mathbb{E}_{(\mathbf{s}_t, \mathbf{a}_t, r_t, \mathbf{s}_{t+1}) \sim \mathcal{D}}\left[\tfrac{1}{2}\left(Q_{\theta_i}(\mathbf{s}_t, \mathbf{a}_t) - (r_t + \gamma[Q_{\theta_i}(\mathbf{s}_{t+1}, \mathbf{a}_{t+1}) - \alpha \log \pi(\mathbf{a}_{t+1}|\mathbf{s}_{t+1})])\right)^2\right] \quad (9)$$

where $\mathcal{D}$ is the experience replay buffer that stores the interactive transition data and is used to train the function approximator, and $i=1,2$ denotes the index of two independent and identical distributed soft-Q functions, respectively.

The policy function that outputs the actions with maximized soft-Q values can also be approximated by policy network $\pi(\phi)$. The objective is calculated as:

$$\mathcal{L}_\pi(\phi) = \mathbb{E}_{\mathbf{s}_t \sim \mathcal{D}}\left[\alpha \log\left(\pi_\phi(f_\phi(\mathbf{s}_t)|\mathbf{s}_t)\right) - Q_{\theta_1}(\mathbf{s}_t, f_\phi(\mathbf{s}_t))\right] \quad (10)$$

where $f_\phi$ denotes a Gaussian distribution derived from the reparameterization trick, and it enables the policy network to become differentiable.

*B. The Optimization Bound of the Inaccurate Model*

Here, we discuss how to design a feasible model-based RL under the inaccurate environment model. Considering the policy training with experience replay buffer, the error produced by the inaccurate model $\epsilon_\mathcal{M}$ can be bounded as:

$$\epsilon_\mathcal{M} = \max_t \mathbb{E}_{\mathbf{s} \sim \pi_t}[D_{TV}(\mathcal{T}(\mathbf{s}_{t+1}|\mathbf{s}_t, \mathbf{a}_t) \| \mathcal{M}(\mathbf{s}_{t+1}|\mathbf{s}_t, \mathbf{a}_t))] \quad (11)$$

where $D_{TV}$ denotes the total variance distance, and $\pi$ denotes the policy to be evaluated.

Since the actions are generated by the ever-updated policy, the deviation between the policy to be evaluated and the policy stored in the experience replay can be bounded by the error $\epsilon_\pi$:

$$\epsilon_\pi \geq D_{TV}(\pi \| \pi_\mathcal{D}) \quad (12)$$

Taking the cumulative reward subjected to a policy in the actual environment as $\eta$ and its counterpart in the constructed virtual environment model as $\eta_\mathcal{M}$, we can achieve the relationship between $\eta$ and $\eta_\mathcal{M}$ within $k$ rollout steps as [28]:

$$\eta^\pi \geq \eta^\pi_\mathcal{M} - 2r_{max} \cdot C(\epsilon_\mathcal{M}, \epsilon_\pi, k) \quad (13)$$

$$C(\epsilon_\mathcal{M}, \epsilon_\pi, k) = \left[\tfrac{\gamma^{k+1}\epsilon_\pi}{(1-\gamma)^2} + \tfrac{\gamma^k \epsilon_\pi}{1-\gamma} + \tfrac{k\epsilon_\mathcal{M}}{1-\gamma}\right] \quad (14)$$

The above formulas indicate that the cumulative reward achieved in the constructed virtual model can approach that of the actual transition with a small region $C$.

Since the error $\epsilon_\pi$ exists inevitably for all off-policy RL algorithms, the other two independent variables become crucial for optimization.

Although the analytical solution of the objective function $C$ is difficult to be found, here we share two useful points. First, it should be noted that the $C$ is non-monotonic with respect to the rollout step but will be increased to infinite with the increasing rollout depth, as the discount factor $\gamma$ is less than 1. Hence, it is not recommended to overuse the virtual environment model, considering its inaccuracy. Second, the model error $\epsilon_\mathcal{M}$ gives rise to monotonically increased $C$, which results in impaired performance. To overcome the above two issues, in this work we develop an adaptive rollout-step adjustor, which is introduced in the following subsection.

*C. Model Uncertainty-aware Rollout Truncation*

The ensemble model constructed in Section III-B provides the ability to evaluate the uncertainty of the environment model. For an arbitrary step, the models are subjected to an uniform distribution $\mathcal{M} \sim \mathcal{U}_\mathcal{M}$. Thus, the mean and variance of the ensemble model can be calculated as:

$$\mu_\mathcal{M}(\mathbf{s}_t, \mathbf{a}_t) = \tfrac{1}{N}\sum_{n=1}^{N} \mu_{\mathcal{M}_\psi^n}(\mathbf{s}_t, \mathbf{a}_t) \quad (15)$$

$$\sigma^2_\mathcal{M}(\mathbf{s}_t, \mathbf{a}_t) = \tfrac{1}{N}\sum_{n=1}^{N} \sigma^2_{\mathcal{M}_\psi^n}(\mathbf{s}_t, \mathbf{a}_t) + \tfrac{1}{N}\left[\sum_{n=1}^{N} \mu^2_{\mathcal{M}_\psi^n}(\mathbf{s}_t, \mathbf{a}_t) - \mu^2_\mathcal{M}(\mathbf{s}_t, \mathbf{a}_t)\right] \quad (16)$$

Note that the above equations only involve the variance, instead of the covariance shown in Eq. (2). The reason is that each action is determined by its paired state, thus, $(\mathbf{s}_t, \mathbf{a}_t)$ represents a single-variable input in practical implementation.

The variance of the ensemble model is set as:

$$\epsilon_\mathcal{M} \propto \sigma^2_\mathcal{M}(\mathbf{s}_t) \quad (17)$$

With this metric, we adopt a truncation method, instead of the conventional one-step or infinity rollout scheme, to determine the prediction horizon. The prospective lengths of rollouts $k^*$ is calculated by a linear operation:

$$k^* = -\omega \cdot \sigma^2_\mathcal{M}(\mathbf{s}_t) + k_{base} \quad (18)$$

where $\omega$ is the weighting factor representing the slope rate, and $k_{base}$ is the baseline value of the expected rollout length.

Combining the above mechanism together with the SAC algorithm in Section IV-A, the procedure of the proposed algorithm can be described in Table I.

TABLE I

### THE UNCERTAINTY-AWARE MODEL-BASED RL ALGORITHM

1: Initialize policy approximator $\pi_\phi$, value approximator $Q_{\theta 1}, Q_{\theta 2}$; initialize experience replay buffer $\mathcal{D}_E$ for storing actual interaction data, $\mathcal{D}_\mathcal{M}$ for storing virtual interaction data.
2: **for** $N_{epoch}$ **do**
3:  Train the virtual environment ensemble model $\mathcal{M}_\psi$ on $\mathcal{D}_E$
4:  **while** not **done**
5:   Execute action according to policy $\pi_\phi$ in actual environment
6:   Obtain next state and reward, store transition tuple to $\mathcal{D}_E$
7:   **for** $M$ model rollouts **do**
8:    Sample $\mathbf{s}_t$ from $\mathcal{D}_E$
9:    Calculate $\sigma^2_\mathcal{M}(\mathbf{s}_t)$ and obtain rollout length $k^*$
10:   **for** $k^*$ rollout steps **do**
11:    Execute action $\pi_\phi(\mathbf{s}_t)$ in random model $\mathcal{M}^i_\psi \sim \mathcal{M}_{ense}$
12:    Store transition tuple to $\mathcal{D}_\mathcal{M}$ and move to next step $\mathbf{s}_t \leftarrow \mathbf{s}_{t+1}$
13:   Update parameters of approximator $\pi_\phi, Q_{\theta 1}, Q_{\theta 2}$ according to the principles of SAC algorithm with the data in the model: $\mathcal{D}_\mathcal{M}$ and $\mathcal{D}_E$
14:   Update the step in the actual environment.

## V. THE APPLICATION TO AUTONOMOUS DRIVING

In this section, the proposed algorithm is further applied to autonomous vehicle to tackle with the end-to-end driving task.

### A. Autonomous Driving Scenarios for Algorithm Validation

The feasibility and effectiveness of the developed algorithm will be validated in the context of autonomous driving through RL's training process and the testing process, respectively. First, the driving scenario for RL training is designed to be a one-way double-lane urban road, wherein three stationary surrounding vehicles and two randomly moving pedestrians are set, as shown in Figure 3(a). With the interaction data generated through the training process, the environment model shall be developed, and then the RL's policy is expected to be further optimized with the virtual interaction data. Thus, the learning efficiency of the algorithm will be evaluated through the training process. And then, the adaptiveness and robustness of the algorithm will be further validated under two additional scenarios shown in Figures 3(a) and 3 (b) in RL's testing phase. All the driving scenarios involved for training and testing are illustrated in details in Fig. 3.

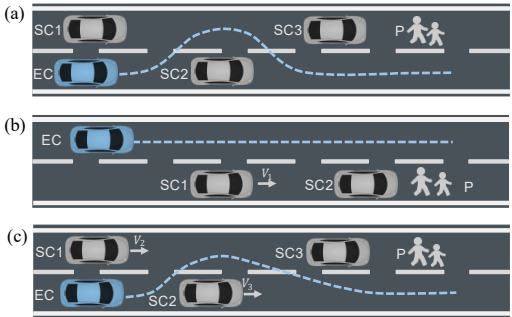

Fig. 3. Schematic diagram of the driving scenarios. SC means surrounding cars, P denotes pedestrians, and EC represents ego car. The blue dotted line depicts one of the feasible trajectories that is expected to be performed by the RL agent. (a): the scenario only for RL training; (b) and (c): the scenarios for RL testing.

In the algorithm validation, we apply the developed new ua-MBRL algorithm to an end-to-end autonomous driving task. Compared to the modular autonomous driving functions, the end-to-end control is more challenging and poses higher requirement on the optimization ability of the algorithm.

### B. Optimization Problem Formulation

To achieve a satisfactory performance, the proposed RL algorithm should be able to drive the ego autonomous vehicle to move smoothly ahead and avoid obstacles and any potential crashes. Following the above goal, the objective function of the optimization problem $\mathcal{J}$ can be formulated as:

$$\mathcal{J} = \omega_1 \cdot C_{colli} + \omega_2 \cdot C_{jerk} + \omega_3 \cdot C_{lane} \quad (19)$$

where $\omega_1$ to $\omega_3$ represents the weighting factors of different sub-objectives, respectively.

$C_{colli}$ quantifies the collision risk by constructing potential-field-like functions, which can be expressed as:

$$C_{colli} = C_{colli,f} + C_{colli,s} \quad (20)$$

$C_{colli,f}$ and $C_{colli,s}$ represent the risks of collision to the frontal and side obstacles, which can be given by:

$$C_{colli,f} = \mathrm{sig}(\|\mathbf{x}_{ego,lon}, \mathbf{x}_{front,lon}\|_2^2) \quad (21a)$$
$$C_{colli,s} = \min_j(\|\mathbf{x}_{ego,lat}, \mathbf{x}_{j,lat}\|_2^2) \quad (21b)$$

where $\mathbf{x}_{ego}, \mathbf{x}_{front}$ denote the position coordinate of the ego vehicle and the frontal obstacle, respectively. The abbreviation $lon$ and $lat$ denote the longitudinal and lateral position, , respectively. The subscript $j$ denotes the side obstacle, i.e., the left and right road-boundaries, as well as surrounding vehicles and pedestrians, if exist. $\mathrm{sig}(\cdot)$ denotes the normalization through sigmoid function, and $\|\cdot\|_2^2$ denotes 2-norm operation.

$C_{jerk}$ is the smoothness constraint by differentiating the yaw rate $\mathbf{\Omega}_{ego}$:

$$C_{jerk} = d\mathbf{\Omega}_{ego}/dt \quad (22)$$

$C_{lane}$ is the expected lane by considering the lateral position of the ego vehicle, which is shown as:

$$C_{lane} = \|\mathbf{x}_{ego,lat}, \mathbf{x}_{tar,lat}\|_2^2 \quad (23)$$

where $\mathbf{x}_{tar,lat}$ denotes the target lateral position.

It should be noted that ideally the safety-associated $C_{colli}$ should serve as a hard constraint in the optimization problem. However, this can hardly be implemented in RL, a gradient-based learning algorithm. Thus, it would be necessary to assign a much greater weight to $C_{colli}$ than to $C_{jerk}$ and $C_{lane}$.

### C. The Reward Shaping

To minimize the objective function developed in Section V-B, the reward $r: \mathcal{S} \times \mathcal{A} \rightarrow \mathbb{R}$ can be established as:

$$r = b - \mathcal{J} \quad (24)$$

where $b$ is the bias of the reward function, set as 0.1.

However, according to the above mechanism, the feedback from the environment is very sparse and can frequently lead the RL agent to local optima. Hence, two shaping techniques, namely, the potential-based scheme [29] and the NGU [30] are employed to improve the reward function.

Specifically, the potential-based reward shaping encourages explorations leveraging heuristic information. In this paper, the prior knowledge is set as the distance to the target coordinate in the longitudinal direction, and the reward term $r_P$ is shown as:

$$r_P = \mathbf{x}_{tar,lon} - \mathbf{x}_{ego,lon} \tag{25}$$

Besides, the NGU shaping technique utilizes a pair of identically initialized deep networks, i.e. the $f(\cdot|\varphi)$ with a fixed weight $\varphi$ and $f(\cdot)$ with an adjustable weight, respectively, to quantify the similarity between the state features and to encourage explorations. The NGU term can be expressed as:

$$r_{NGU} = \min\left(\left[\frac{\|f(\mathbf{s}_{t+1}|\varphi) - f(\mathbf{s}_{t+1})\| - \mathbb{E}[f(\mathbf{s}_{t+1}|\varphi)]}{\sigma[f(\mathbf{s}_{t+1}|\varphi)]}\right]^+ + 1,\ L\right) \tag{26}$$

where the $\|\cdot\|$ denotes the 1-norm, calculating the similarity between the two networks. $[\cdot]^+$ denotes that outputting the original result when the value is positive, or outputting zero when the value is negative. $\sigma$ denotes the standard deviation, and $L$ is a regularization term.

Since the parameters of the adjustable network $f(\cdot)$ are trained with the state variables, the difference between its output and that of the fixed $f(\cdot|\mathbf{s}_{t+1})$ becomes smaller when the RL agent revisits those trained states. Thus, the NGU reward term encourages explorations regarding unseen states. The network structure and parameters of the NGU reward scheme are provided in Table II.

Using the above two shaping techniques, the ultimate reward function can be re-written as:

$$r = b - \mathcal{J} + r_P + r_{NGU} \tag{27}$$

TABLE II
DETAILS OF THE NGU REWARD SCHEME

| Parameter | Value |
|---|---|
| Input shape | 64×64×1 |
| Fixed net convolution features | [8,16] (kernel size 5×5, stride 2) |
| Fixed net dense features | [128] |
| Adjustable net convolution features | [8,16] (kernel size 5×5, stride 2) |
| Adjustable net dense features | [128] |

*D. MBRL-based Autonomous Driving Algorithm*

To formulate a completed RL problem under autonomous driving task, the state space and action space need to be defined.

Although there are multi-modal onboard sensors available for perception in AV applications, in this work we only utilize the frontal-view output provided by a semantic camera as the state variable. By compressing the original image into 64×64-pixel matrix with reshaping, the state space can be expressed as:

$$\mathcal{S} = \{\mathbf{p}|\mathbf{p} \in [0,255]\}_{1\times 4096} \tag{28}$$

where $\mathbf{p}$ denotes the value of the 8-bit pixel.

In addition, the RL agent only takes charge of the lateral control task of the AV. Thus the action space can be given by:

$$\mathcal{A} = \{\boldsymbol{\delta}|\boldsymbol{\delta} \in [-\pi/2, \pi/2]\} \tag{29}$$

where $\boldsymbol{\delta}$ denotes the steering wheel angle. A negative value indicating turning left, while a positive one refers to turning right. The range of the steering wheel manipulation is set to be adaptive to the scenario requirement.

The vehicle's longitudinal dynamics is controlled by a proportional-integral controller, which can be expressed as:

$$\mathbf{acc} = K_P(\mathbf{v}_{ego,lon} - \mathbf{v}_{tar,lon}) + K_I \int (\mathbf{v}_{ego,lon} - \mathbf{v}_{tar,lon}) \tag{30}$$

where $\mathbf{v}$ and $\mathbf{acc}$ denote the longitudinal velocity and the acceleration, respectively. $K_P$ and $K_I$ denote the proportional and integral coefficients of the controller, respectively.

The hyperparameters of the RL algorithm are provided in Table III. We utilize neural networks to construct the RL approximators. All the neural networks involved, including the virtual environment model, and the actor and critic functions of the RL, are trained with the Adam optimizer. The details are provided in the Appendix Table I-III.

TABLE III
HYPERPARAMETERS OF THE MBRL-BASED AV STRATEGY

| Symbol | Meaning | Value |
|---|---|---|
| $\mathcal{C}_{\mathcal{D}_\mathcal{M}}$ | Buffer size of model experience | 2e15 |
| $\mathcal{B}_\mathcal{M}$ | Batch size of model training | 128 |
| $l_\mathcal{M}$ | Initial learning rate of model training | 0.0005 |
| $\mathcal{C}_{\mathcal{D}_E}$ | Buffer size of practical experience | 2e15 |
| $\mathcal{B}_{RL}$ | Batch size of RL training | 64 |
| $l_r$ | Initial learning rate of RL training | 0.0001 |
| $N_{epoch}$ | Maximum episode of RL training | 500 |
| $\alpha$ | Temperature coefficient of RL training | 0.005 |
| $\gamma$ | Discount factor of Bellman equation | 0.97 |
| $\tau$ | Soft updating factor of target networks | 0.005 |

## VI. EXPERIMENTAL RESULTS AND DISCUSSIONS

*A. Experimental Set-up*

The validation of the proposed approach is divided into two stages, namely, the validation in RL's training phase and the validation in RL's testing phase. In addition, more discussions and analysis on the algorithm's mechanism are also provided. The three scenarios depicted in Fig. 3 are all employed to construct the controlled environment in validation. Specifically, Scenario (a) is only used in the training phase, while all three scenarios are utilized in the testing phase.

To achieve a high-fidelity validation, CARLA simulation platform is used to develop the driving environment [31]. The simulations are conducted on a computer with a CPU of intel i9-9900K and a graphic card of RTX 2080. The frequency for rendering animation is set as 50 Hz, and the control command frequency is 20 Hz. All associated learning algorithms and scripts are programmed by python.

*B. Algorithm Validation in the Training Process*

There are two tasks of the algorithm validation in RL's training process. First, the performance of the trained environment model is evaluated, as its quality is crucial for the subsequently developed MBRL algorithm. Then, we evaluate the training efficiency and performance of the developed ua-MBRL method.

As illustrated in Fig. 4, multiple indicators are adopted to evaluate the prediction accuracy of the environment model. For a typical ensemble model, the training loss measured by the mean squared error (MSE) is provided in Fig. 4(a). The gradually declined losses for both the predicted state and reward suggest the model accuracy is improved. In Fig. 4(b), the standard deviations of the outputs of ensemble models are calculated. The declining curve indicates that the independently distributed models in the set could gradually converge to a consistent level. Hence, it is validated that the prediction confidence of the developed model is effectively improved, and the model uncertainty is reduced accordingly throughout the training process.

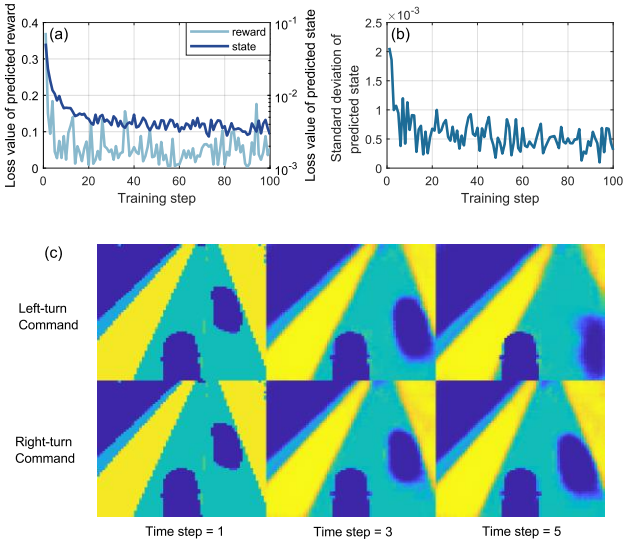

Fig.4. Prediction overview of the parameterized environment model. (a), the loss value of predicted reward and state, (b), standard deviation of predicted state variable of ensemble models, (c), a typical demonstration of action-conditioned rollout prediction processes starting from the same practical state. In the semantic image, the bottom-middle vehicle is the ego car, while one stationary adjacent vehicle locates on the right side. The three consecutive images in the top row reveal a predicted state transition process under continual and identical left-turn command, and the images in the below row correspond to the situation under right-turn command.

Further, we visualize a typical action-conditioned prediction process in Fig. 4 (c) to illustrate the model's effectiveness. The ego vehicle is initially located at the bottom medium in the image, and one adjacent vehicle is on the right side. Starting from this state, we give different actions to the environment model and observe their rollout predictions. The prediction horizon is set as five steps, and three sequential images are presented in this process. It is noticed that the environment model succeeds to predict the longitudinal driving behavior of the ego vehicle under both input actions, as witnessed by the backward-moving adjacent vehicle in the images. While, more importantly, the model conducts explicitly distinct lateral behaviors when receiving different actions. The left-turn command leads to an out-of-bound situation, while the right-turn instruction causes a rightward moving. This observation suggests that our model can generally generate correct predictions, indicating its feasibility. Nevertheless, there inevitably exists some inaccurate predictions, for instance, the blurred shape of the surrounding vehicle, and the inconsistent driving distances of the ego vehicle under left-turn and right-turn behaviors. These inaccuracies and uncertainties of the prediction would affect the performance of the subsequent learning processes of the RL agent.

To assess the training performance of the RL agent in the autonomous driving task, the episodic average reward is evaluated. Specifically, two metrics are adopted for the evaluation. The learning efficiency, which is indicated by the number of episodes when reaching the same level of reward, and the asymptotic performance, which is obtained from the approximately converged reward curve. Some state-of-the-art methods are employed as benchmarks for comparison. First, we use the model-assisted bootstrapped (MAB) approach [23] to establish the instantiation on the SAC algorithm. The formed MAB-SAC, as a baseline method, can well describe the capability of the current Dyna-style MBRLs. Second, we establish another baseline based on the model-based value expansion (MVE) approach [32]. The formed MVE-SAC serves as a representative candidate of the value-augmentation MBRL scheme. Lastly, the Vanilla-SAC [33] is utilized to provide a benchmark of the model-free scheme. For each method, multiple trials are conducted with different random seeds to reflect their average performance. The comparison results of different RL algorithms are illustrated in Fig. 5. The solid lines represent the mean values of the training curves, and the error bands are 1-$\sigma$ intervals.

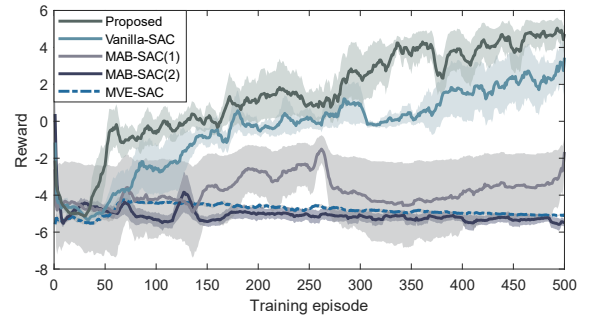

Fig.5. Training rewards of the proposed algorithm and baselines. Curves are flattened. Two schemes of MAB-SAC are evaluated: (1) and (2) denote the utilization of environment model's prediction with rollouts of 5-step and 10-step, respectively. MVE-SAC uses a 10-step prediction of the environment model.

It is observed that the proposed algorithm is superior over all baselines. Taking the ultimate reward of Vanilla-SAC as the basis, the proposed algorithm achieves an improvement of 42.8% in learning efficiency. However, two model-based baselines, MAB-SAC and MVE-SAC, fail to achieve a comparable asymptotic performance, even compared to the Vanilla-SAC. More specifically, the MAB-SAC with a shorter prediction horizon obtains a slightly lower training reward than the Vanilla-SAC, while its counterpart with a longer prediction horizon, as well as the MVE-SAC, fail to improve their policies throughout the training course.

Behind the above observations, the reason causes the unfavorable performance of the baseline methods is the model inaccuracy. The critic's evaluation regarding the optimal policy essentially determines the learning ability of RL. However, when significant errors accumulate in the prediction rollouts, the model-based baseline algorithms receive over unrealistic samples of state variables. Such prediction data introduces severe biases to the data distribution and impairs the learning ability. Since the complex driving environment inevitably leads to poor model predictions, model-based baselines, like MAB-SAC and MVE-SAC, can hardly improve autonomous driving performance within the presented configuration. Evidence supporting this idea can be found that the MAB-SAC with a

> REPLACE THIS LINE WITH YOUR PAPER IDENTIFICATION NUMBER (DOUBLE-CLICK HERE TO EDIT) <    8shorter horizon achieves a remarkably better performance than the one with a long prediction horizon. However, the proposed algorithm determines the familiarity of each state to evaluate its confidence in executing corresponding predictions in real-time. Consequently, truncating the unrealistic prediction mitigates the misleading evaluation of the critic function, and therefore effectively improves the learning capability.

Overall, the algorithm validation through the training process highlights the advance of the proposed algorithm. The comparison results shown in Fig. 5 indicates the feasibility and effectiveness of the ensemble-based uncertainty evaluation and rollout truncation in dealing with model uncertainty.

### C. Algorithm Validation in the Testing Process

In addition to the above validation on the training efficiency, the adaptiveness, robustness and implementation performance of the proposed algorithm needs to be further evaluated in RL's testing process. In this phase, the policies generated by different RL algorithms will be tested across all three designed driving scenarios. Besides the aforementioned RL-related baselines, the imitation learning (IL) method [34] is also introduced as a state-of-the-art supervised learning to provide a more comprehensive comparison of learning-based policies for autonomous driving. For the IL method, we utilize a PI controller with prior knowledge to generate the control sequences corresponding to the state variables. The generated input-output data is then fed to the IL agent to enable the learning from demonstration.

First, we test the adaptiveness of different methods. The metric of the episode duration, representing the completed driving distance until the task ending or collision happening, is adopted. The tests are repeated for multiple rounds, and the metric is represented with the mean value and the standard error.

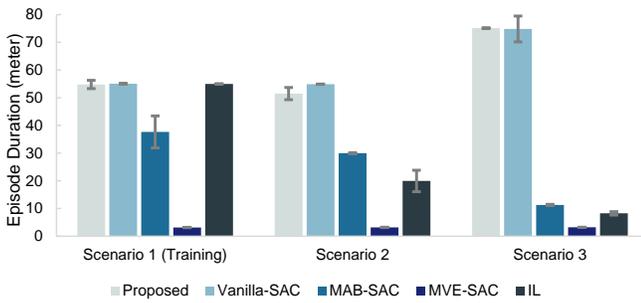

Fig.6. Results of the episode duration of the proposed algorithm and baselines in different testing scenarios.

Based on the results shown in Fig. 6, it is observed that the policies generated by both the proposed algorithm and the model-free Vanilla-SAC can perform well and achieve a equally high level against the IL in Scenario 1, the trained scenario. While in untrained scenarios, the performance of the imitation learning policy drops remarkably, while the proposed algorithm and the Vanilla-SAC are able to remain at a high level. This suggests that RL-based policies have better adaptiveness in the designed testing tasks for autonomous driving. When comparing only among RL-based algorithms, the performances of the proposed algorithm and the Vanilla-SAC are comparably good. It is particularly satisfactory given the fact that the MBRL algorithms are trained with inaccurate prediction data. However, other state-of-the-art model-based RL baselines, i.e., the MAB-SAC and MVE-SAC, struggle to control the ego vehicle towards success in the tasks. The above results again demonstrates the superiority of the proposed method comparing to other existing MBRL methods, in terms of the practical implementation performance.

Next, we evaluate the robustness of the all the obtained learning policies. Specifically, we re-conduct the experiments of the above adaptiveness testing but impose an additional small-scale Gaussian noise (10% with respect to the control domain) to the policy outputs. Then, the episode duration could be possibly undermined due to the noise injection. The new testing results against the noise are compared to the non-noise injection tests (Fig. 6) using confrontational bar plots in Fig. 7.

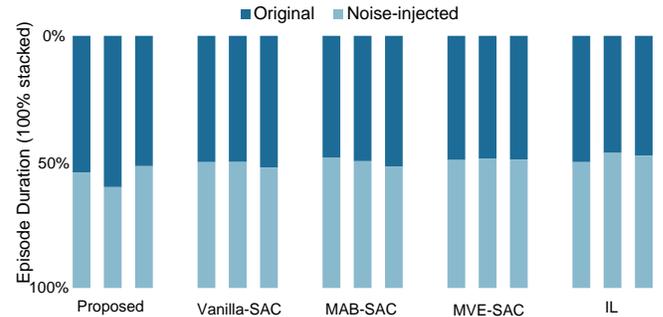

Fig.7. Robustness testing of the proposed algorithm and baselines in different autonomous driving scenarios. The three bars in one cluster represent the three testing scenarios defined in Fig.3.

According to the results shown in Fig. 7, the roughly even-divided bars indicate that RL-based policies achieve favorable robustness in the autonomous driving tasks. It should be noted that this anti-noise ability is derived from the robust learning characteristics of the maximum entropy RL algorithm. Our proposed method reserves the merit of robustness from the Vanilla-SAC algorithm, showing its great potential in the applications to autonomous driving.

Lastly, the driving smoothness is evaluated during the implementation by using vehicle dynamics states, including the yaw rate and lateral velocity. To this end, we design the driving task to be conducted in a flat road without surrounding obstacles, and a straight trajectory is expected to be performed. Therefore, neutral indicators, i.e. the zero lateral velocity and zero yaw rate, can be taken as the central point of the average absolute deviation (AAD). And the AAD statistical results are adopted as the key metric, and a smaller AAD indicates a better driving smoothness.

The results shown in Fig. 8(a) suggest that all the three involved policies are able to perform approximately straight driving. Nonetheless, all the policies perform results in deviations to the neutral line to different extents. The reason behind is that all policies are trained with the lane-change scenario depicted in Fig. 3(a) only. But among the three methods, the two RL-based policies have better adaptiveness. The results illustrated in Figs. 8(b)-(c) suggest that the performances of the policies derived from the proposed algorithm and the Vanilla-SAC are comparable. More specifically, in terms of the lateral velocity, the proposed algorithm achieves an AAD of 0.2501, which is slightly worse than that of the Vanilla-SAC (0.2494). And the similar trend is witnessed regarding the results of yaw rate. The AAD values of the proposed algorithm and the Vanilla-SAC are 0.0548 and 0.0542, respectively. It should be noted that under the task of end-to-end driving, the policies would inevitably result in fluctuations on lateral control. This problem could be mitigated if we switch the end-to-end paradigm to the modular autonomous driving approach, where the RL-based policy can be used only for decision-making.

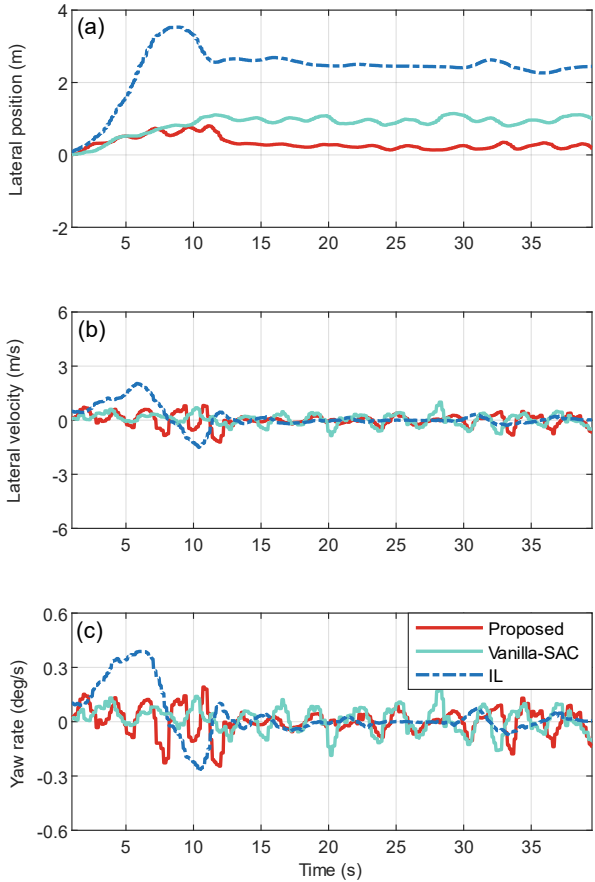

Fig.8. The smoothness testing on the straight-line driving scenario without surrounding obstacles. (a) the lateral position of the ego vehicle, (b) the lateral velocity of the ego vehicle, (c) the yaw rate of the ego vehicle. In all three subplots, the positive value in y-axis represents the leftward displacement relative to the initial position, while the negative value represents the rightward displacement.

### D. Discussion and Analysis on the Algorithm Mechanism

In this section, we further investigate the essence and mechanism of the proposed algorithm. First, the functionality of the algorithm is examined by presenting various truncating steps. Then, the effectiveness of the algorithm is evaluated by comparing the proposed scheme to those fixed-truncating-step schemes.

To visualize the adjustment of the truncation, the state at the spawn point of the ego vehicle is particularly observed, as it will be visited in each episode. The derived prediction of this initial state is recorded throughout the training process and is plotted in Fig.9. The significantly varying prediction horizon indicates the effective adjustment of the truncating step. Nonetheless, the prediction capability increases and converges to a specified value (6 steps) maximum along with the process, as witnessed by the overall trend of the scatter density.

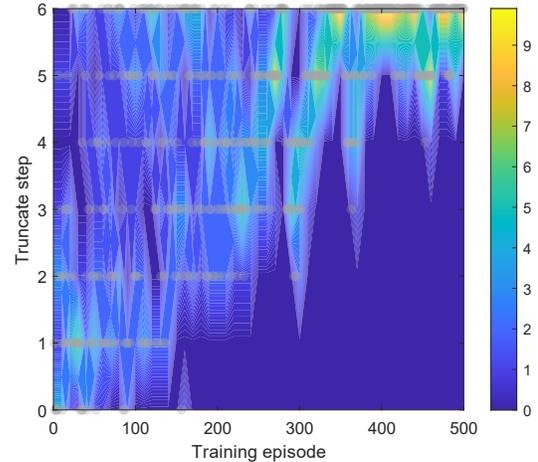

Fig.9. Results of the truncating-step value of one representative state during the training process. At the first visit of the chosen state of each episode, the rollout horizon is determined according to the assessed model uncertainty and recorded. The gray points are the actual truncating points, at which further prediction is terminated. The density map is calculated by the scatter distribution.

In addition, an ablation study on the effect of the truncation is conducted, as shown in Fig. 10. Within the proposed ua-MBRL architecture, the fixed-step candidates fail to compete with the proposed truncation-based algorithm, with respect to the learning efficiency. And at a same level of the reward, among all the fixed-step schemes, only the one with 1-step truncation performs better than the model-free Vanilla-SAC. The results suggest that the proposed model uncertainty-based truncation is decisive to the algorithm performance. Apart from the proposed method, an interesting finding is that the longer the prediction horizon is, the worse performance the algorithm achieves. Considering that the long prediction horizon induces uncertainty and inaccuracy, this finding once again validates the fundamental issue of the model inaccuracy laying in the MBRL methods.

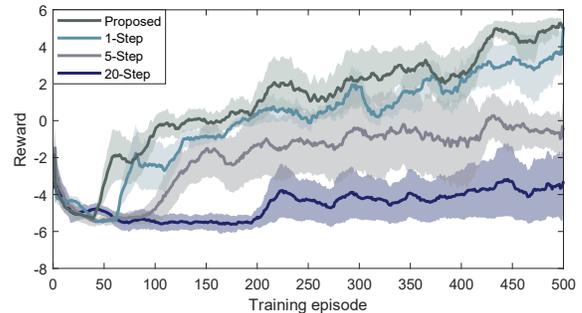

Fig.10. Results of the training reward of the proposed algorithm and the fix-step baselines. Curves are flattened. Baselines adopt the same MBRL architecture but use fix-step prediction instead of the truncation-based one.

Overall, the proposed algorithm provides an effective solution to further improving MBRL, and it is particularly promising in the applications with the environment with modeling uncertainty.

## VII. CONCLUSION

In this paper, a novel uncertainty-aware model-based reinforcement learning method is proposed and validated in end-to-end autonomous driving tasks. An action-conditioned environment model is established to map the transition dynamics of the RL agent in autonomous driving scenarios. Leveraging the prediction rollouts provided by the model, the algorithm adopts the generated virtual experience when it is reliable and truncates the prediction when encountering high uncertainty of the model. The proposed algorithm is then applied to end-to-end autonomous driving under various scenarios, evaluating its learning performance. Based on the validation results, three major conclusions can be drawn:

1) The feasibility and effectiveness of the proposed ua-MBRL method are validated under the autonomous driving tasks with model uncertainty. Its performance is advantageous over other existing model-based RL approaches under the designed testing scenarios.

2) The proposed UA-MBRL algorithm outperforms the state-of-the-art model-free RL baseline method, the soft actor critic, by 42.8%, with respect to the learning efficiency.

3) The policy obtained by the proposed UA-MBRL achieves good performance under various tasks in autonomous driving, demonstrating its adaptiveness and robustness.

## APPENDIX

TABLE I
DETAILS OF THE NETWORK FOR THE ENVIRONMENT MODEL

| Parameter | Value |
| --- | --- |
| Input Shape (state + action) | $4099 \times 1 + 1 \times 1$ |
| Convolution Channels (to latent) | [8,16,32,64] |
| Convolution Kernels | 5×5, 5×5, 5×5, 5×5 |
| Convolution Strides | 2, 2, 2, 1 |
| Deconvolution Channels (to next image) | [64,32,16,8] |
| Deconvolution Kernels | 4×4, 5×5, 6×6, 6×6 |
| Deconvolution Strides | 1, 2, 2, 2 |
| Physical Embedding | [512] |
| Action Embedding | [512] |
| Dense Feature (to next physical state) | [3] |
| Dense Feature (to reward) | [512,128,1] |

TABLE II
DETAILS OF THE CRITIC (TARGET) NETWORKS

| Parameter | Value |
|---|---|
| Input Shape (state + action) | 64×64×1 + 1×1 |
| Convolution Features | [8,16,32], kernel 6×6, Stride=2 |
| Dense Feature | [256, 64, 64, 1] |
| Action Embedding (to 2nd Dense) | [64] |

TABLE III
DETAILS OF THE ACTOR (TARGET) NETWORKS

| Parameter | Value |
|---|---|
| Input Shape (state) | 64×64×1 |
| Convolution Features | [8,16,32], kernel 6×6, Stride=2 |
| Dense Feature | [256, 64, 1] |